\documentclass[runningheads]{llncs}
\usepackage[T1]{fontenc}
\usepackage{graphicx,bbding,amsmath,amssymb,amsfonts,cleveref,bm,multirow,booktabs,textcomp,algorithmic,tikz,tikz-qtree,subcaption,framed,tabularx,adjustbox,arydshln,enumitem}
\begin{document}
\title{Generative Sentiment Analysis via Latent Category Distribution and Constrained Decoding}
%
\titlerunning{Generative Sentiment Analysis via LCD and CD}
%

\author{Jun Zhou\inst{1} \and
    Dongyang Yu\inst{2} \and
    Kamran Aziz \inst{1} \and
    Fangfang Su\inst{3} \and
    Qing Zhang \inst{4} \and
    Fei Li \inst{1} \and
    Donghong Ji \inst{1(}\Envelope\inst{)}
    }
\authorrunning{Jun Zhou et al.}
%
\institute{
    Key Laboratory of Aerospace Information Security and Trusted Computing, Ministry of Education, School of Cyber Science and Engineering, Wuhan University, China 
    \and Beijing Rigour Technology Co., Ltd., China
    \and School of Cyberspace, Hangzhou Dianzi University, China
    \and North China Power Engineering Co., Ltd. of China Power Engineering Consulting Group, China\\
    \email{dhji@whu.edu.cn}
    \\
    }

%
\authorrunning{J. Zhou et al.}
%

%
\maketitle              
%

\begin{abstract}
Fine-grained sentiment analysis involves extracting and organizing sentiment elements from textual data. However, existing approaches often overlook issues of category semantic inclusion and overlap, as well as inherent structural patterns within the target sequence. This study introduces a generative sentiment analysis model. To address the challenges related to category semantic inclusion and overlap, a latent category distribution variable is introduced. By reconstructing the input of a variational autoencoder, the model learns the intensity of the relationship between categories and text, thereby improving sequence generation. Additionally, a trie data structure and constrained decoding strategy are utilized to exploit structural patterns, which in turn reduces the search space and regularizes the generation process. Experimental results on the Restaurant-ACOS and Laptop-ACOS datasets demonstrate a significant performance improvement compared to baseline models. Ablation experiments further confirm the effectiveness of latent category distribution and constrained decoding strategy.

\end{abstract}
%
%
%
%
\section{Introduction}
\label{section: Introduction}
Sentiment analysis, especially the fine-grained Aspect-Based Sentiment Analysis (ABSA), plays a vital role in Natural Language Processing (NLP) \cite{pang2008opinion}. ABSA aims to convert unstructured text into structured records to achieve a profound comprehension of sentiment semantics.
For instance, in the comment ``\textit{as far as gaming performance, the m370x does quite well}'', the ABSA quadruple extraction task \cite{CaiXY20} seeks to extract aspect-level sentiment elements, including \textit{aspect}, \textit{category}, \textit{opinion}, and \textit{sentiment} polarity, represented as a structured quadruple $\langle$\textit{m370x}, \textit{GRAPHICS}\textit{\#}\textit{OPERATION}\textit{\_}\textit{PERFORMANCE}, \textit{well}, \textit{POS}$\rangle$. 
Given its ability to offer opinions and sentiments regarding specific aspects, this task has been widely used in domains such as marketing and conversational agents \cite{pang2008opinion}.

Sentiment extraction models are currently categorized into two main groups: extractive methods and generative methods \cite{Zhang0DBL20}. Extractive methods typically frame the task as a sequence labeling or multi-classification problem \cite{CaiXY20}, necessitating task-specific classification networks \cite{MaoSYZC22}. 
The introduction of generative pre-trained language models such as T5 \cite{RaffelSRLNMZLL20} and BART \cite{lewis-etal-2020-bart} has introduced a new paradigm for information extraction, shifting tasks towards sequence generation \cite{ZhangD0YBL21}. This generative approach has also demonstrated promising results in sentiment quadruple extraction tasks \cite{ZhangD0YBL21}.


However, existing sentiment extraction models often overlook the challenges posed by category inclusion and overlap, which refer to the semantic inclusion or overlapping of categories within the dataset.
For example, the category ``\textit{GRAPHICS}'' is semantically encompassed by categories such as ``\textit{HARDWARE}'' and ``\textit{LAPTOP}'' (where ``\textit{GRAPHICS}'' is a semantic subset of ``\textit{HARDWARE}'', and ``\textit{HARDWARE}'' is a semantic subset of ``\textit{LAPTOP}''). Similarly, categories like ``\textit{MEMORY}'', ``\textit{KEYBOARD}'', and ``\textit{BATTERY}'' exhibit similar semantic inclusion relationships with ``\textit{HARDWARE}'' and ``\textit{LAPTOP}''. Additionally, ``\textit{GRAPHICS}'' and ``\textit{DISPLAY}'' share a semantic overlap, as do ``\textit{WARRANTY}'' and ``\textit{SUPPORT}''. 
These instances of semantic inclusion and overlap complicate the accurate extraction of elements, a concern often neglected by current models.

On the other hand, current generative models typically employ greedy decoding \cite{0001LXHTL0LC20}, selecting the word with the highest probability from the entire target vocabulary at each time step. Nonetheless, this strategy fails to effectively utilize pattern knowledge inherent in the target sequence, which could otherwise constrain the search space \cite{0001LXHTL0LC20}. To elaborate, datasets typically define restricted sets of categories, subcategories, and sentiment polarities. Greedy decoding based on the entire vocabulary may lead to the generation of invalid categories, misaligned subcategories, as well as invalid sentiment polarities. Despite these challenges, current generative models often neglect the significance of pattern knowledge in the target sequence.



Motivated by recent advances in generative frameworks, this study proposes a T5-based model for sentiment quadruple extraction. The model treats extraction as a generative process, producing sentiment elements in a uniform, natural language format. 
To address category semantic inclusion and overlap issues, a Latent Category Distribution (LCD) variable is introduced. 
This variable, learned through reconstructing the input of a Variational AutoEncoder (VAE) \cite{fei2020latent}, captures the relationship intensity between text and categories, supplementing target sequence generation.
Furthermore, to leverage the inherent pattern knowledge within the target sequence, a Constrained Decoding (CD) strategy based on a trie data structure \cite{ChenBHK20} is adopted. 
This strategy introduces structural pattern knowledge during the decoding stage to reduce the search space and enhance sequence generation.
Evaluation on the Restaurant-ACOS \cite{CaiXY20} and Laptop-ACOS \cite{CaiXY20} datasets show that the proposed model outperforms baseline models, and demonstrate the efficacy of the LCD and CD.

The primary contributions of this study include:
\begin{itemize}
	\item Addressing category semantic inclusion and overlap challenges via a latent category distribution variable learning the relationship intensity between text and categories.
	\item Regulating the generation process through a constrained decoding strategy leveraging structural pattern knowledge within the target sequence.
	\item Experimental results on the Restaurant-ACOS and Laptop-ACOS datasets demonstrate the effectiveness of the proposed model.
\end{itemize}

\section{Related Works}
\label{section: Related Works}

In recent years, there has been a growing interest in fine-grained sentiment analysis \cite{pang2008opinion}, involving the extraction and organization of sentiment-related elements such as \textit{aspects}, \textit{categories}, \textit{opinions}, and \textit{sentiment} polarities from text. Initially focusing on individual elements such as aspect terms \cite{pang2008opinion}, categories \cite{BuRZYWZW21}, and sentiment polarities \cite{ZhangQ20}. Early research has expanded to include paired extractions \cite{zhao-etal-2020-spanmlt}. Subsequent studies have advanced the field by proposing tasks like triplet extraction \cite{peng2020knowing}, involving aspects, opinions, and sentiment polarities. More recently, quadruple extraction \cite{CaiXY20} has been introduced to extract aspects, categories, opinions, and sentiment polarities concurrently, including implicit elements \cite{CaiXY20}. This task has garnered significant attention in the research community.

Quadruple extraction models can be broadly categorized into two main approaches \cite{ZhangD0YBL21}: extractive and generative methods. Extractive methods decompose the task into distinct subtasks, addressing them separately \cite{ZhangD0YBL21}. For instance, Cai et al. \cite{CaiXY20} employ either the Double Propagation (DP) \cite{QiuLBC11} algorithm or the JET \cite{xu2020position} model to extract \textit{aspect}-\textit{opinion}-\textit{sentiment} triplets, and then predict the category of these triplets. Alternatively, they might use the TAS-BERT \cite{wan2020target} model for the joint extraction of \textit{aspect}-\textit{opinion} pairs, followed by category and sentiment polarity prediction. Zhu et al. \cite{ZhuBXLZK23} capture local and contextual features and then utilize a grid labeling scheme along with its decoding method to extract quadruples. 
Extractive methods often transform the task into sequence labeling or multi-classification problems, requiring the development of task-specific classification networks \cite{MaoSYZC22}.


The emergence of generative language models \cite{lewis-etal-2020-bart,RaffelSRLNMZLL20} has introduced a novel paradigm to information extraction by unifying tasks into sequence generation problems, eliminating the need for task-specific network designs. 
This shift simplifies model design and facilitates the comprehensive use of correlation information among different subtask modules \cite{MaoSYZC22}. 
Generative extraction has been applied to various tasks \cite{0001LXHTL0LC20}. Efforts have recently also focused on transforming fine-grained sentiment analysis tasks into sequence generation challenges. For example, the BARTABSA \cite{yan-etal-2021-unified} model reformulates sentiment element extraction as a problem of generating indices for target elements. The Paragraph \cite{ZhangD0YBL21} model employs predefined templates to reframe quadruple extraction as a template-based generation problem. The GAS \cite{Zhang0DBL20} model proposes annotation-based and extraction-based approaches to model the task as a text generation problem.


However, the models discussed overlook challenges posed by category semantic overlap and inclusion, and neglect to utilize structural pattern knowledge to constrain decoding. This study, inspired by the generative paradigm, adopts a generative model as its core framework. Unlike existing generative models, it addresses semantic overlap and inclusion by incorporating a latent category distribution, and utilizes pattern knowledge to guide decoding during generation. Experimental results validate the effectiveness of the proposed model.


\section{Task Definition}
\label{section: Task Definition}

The objective of sentiment quadruple extraction is to extract structured quadruples $\langle a, c, o, s \rangle $ from a given comment $ X = [x_1, \ldots, x_N] $ \cite{CaiXY20}:
\begin{equation}
	X \Rightarrow [\langle a_1, c_1, o_1, s_1 \rangle, \langle a_2, c_2, o_2, s_2 \rangle, \ldots]
\end{equation}
where $a$ represents the \textit{aspect}, $c$ denotes the \textit{category} to which the aspect belongs, $o$ signifies the \textit{opinion}, and $s$ indicates the \textit{sentiment} polarity of the opinion. 

Since this study adopts a generative manner to the extraction task, using a Sequence-to-Sequence (Seq2Seq) framework to convert $X$ into a linearized augmented language sequence $Y$ \cite{0001LXHTL0LC20}, the task is specifically transformed into:
\begin{equation}
	X \Rightarrow Y = [y_1, y_2, \ldots] 
\end{equation}
where $Y$ is a word sequence consisting of the essential elements of the quadruples, which can be extracted by parsing $Y$.


\section{Model Structure}
\label{section: Model Structure}
\begin{figure}[htbp]
	\includegraphics[scale=0.5]{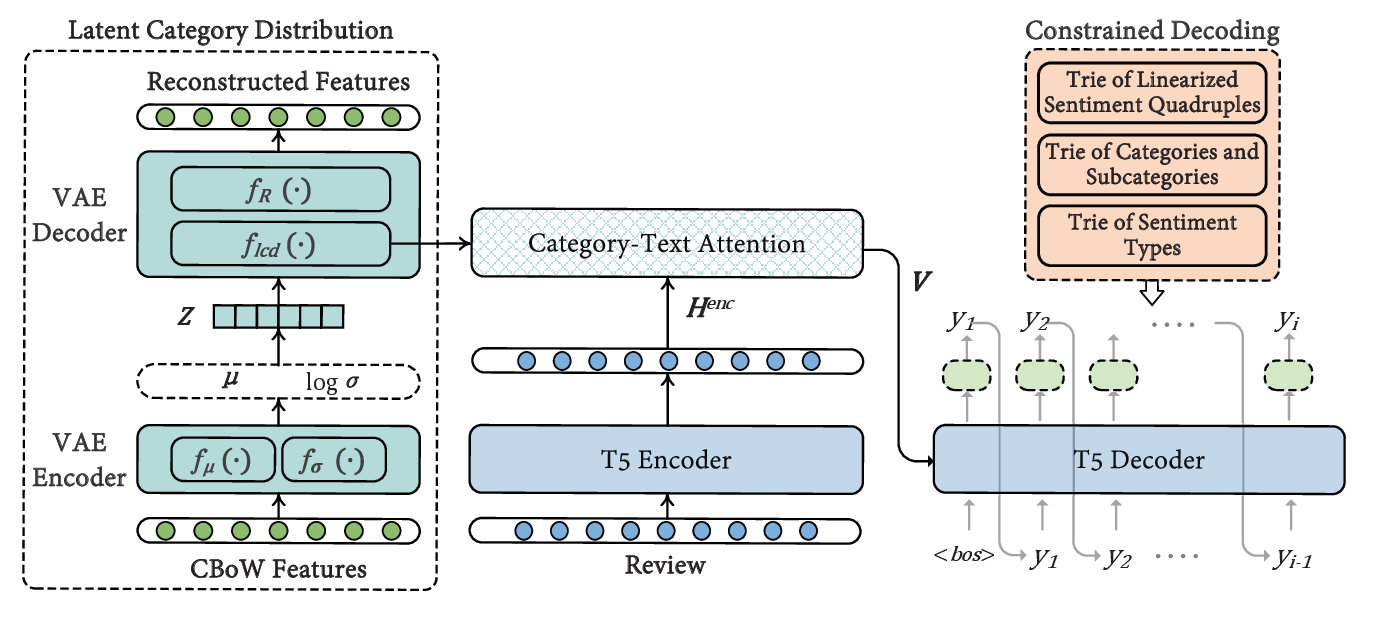}
	\caption{Model architecture}
	\label{fig: architecture}
\end{figure}

\subsection{Latent Category Distribution}
\label{section: Latent Category Distribution}

Latent Category Distribution (LCD) module (refer to \Cref{fig: architecture}) models category distribution within text following Fei et al. \cite{fei2020latent}, using a latent variable instead of explicit measurement. It reconstructs the variational autoencoder input to learn the LCD representation $\bm{Z}$.

\subsubsection{Encoder}

The input to the LCD module is the Category Bag of Words (CBoW) features \cite{fei2020latent} $\bm{X}_{CBoW} \in \mathbb{R}^L$. The construction of CBoW features involves: 
1) Removing stopwords, meaningless words, and sentiment-related words from texts while preserving category- and aspect-related words to construct a CBoW vocabulary of length $L$. This step aims to ensure that representation $\bm{Z}$ contains comprehensive category information rather than general semantic information, while decreasing the BoW vocabulary size is beneficial for VAE training \cite{fei2020latent}. 
2) Calculating the occurrence frequency of words in the vocabulary appearing in a sentence to obtain the sentence's CBoW features, forming a $L$-dimensional vector. 


The encoder of the LCD module comprises two Feedforward Neural Networks (FFNs), denoted as $f_{\mu}(\cdot)$ and $f_{\sigma}(\cdot)$, respectively. Each of these networks includes a hidden layer with a dimensionality of 256, which is responsible for transforming the features $\bm{X}_{CBoW}$ into the prior parameters $\bm{\mu}$ and $\bm{\sigma}$:
\begin{equation}
	\bm{\mu} = f_{\mu} ( \bm{X}_{CBoW} )
\end{equation}
\begin{equation}
	\text{log}\bm{\sigma} = f_{\sigma} ( \bm{X}_{CBoW} )
\end{equation}
Following Bowman et al. \cite{BowmanVVDJB16}, a latent variable $\bm{Z}' = \bm{\mu} + \bm{\sigma} \cdot \epsilon \in \mathbb{R}^K$ is defined to represent the latent category distribution, where $K$ is the number of categories, $\epsilon$ is a Gaussian noise variable sampled from $\mathcal{N}(0,1)$. $\bm{Z}'$ is then normalized as follows:
\begin{equation}
	\bm{Z} = \text{softmax}(\bm{Z}')
\end{equation}
Consequently, the latent category distribution can be reflected by variable $\bm{Z}$.
\subsubsection{Decoder}
The LCD module employs variational inference \cite{MiaoYB16} to approximate a posterior distribution over $\bm{Z}$. It begins by applying a linear transformation $f_{lcd}(\cdot)$ to $\bm{Z}$:

\begin{equation}
	\bm{R}^{lcd} = f_{lcd}(\bm{Z}, \bm{W}_{lcd})
\end{equation}
where $\bm{W}_{lcd} \in \mathbb{R}^{K \times dim}$, $dim$ is the encoding dimension. $\bm{R}^{lcd}$ will be fed into the generative framework (\Cref{subsection: Generative Framework}) with category distribution information to enhance the  generation.
Subsequently, a FFNs $f_R(\cdot)$ is employed to further decode $\bm{R}^{lcd}$ into $\bm{R}$:
\begin{equation}
	\bm{R} = f_R(\bm{R}^{lcd}, \bm{W}_R)
\end{equation}
where $\bm{W}_R \in \mathbb{R}^{L \times dim}$.
Finally, the CBoW features are reconstructed by:
\begin{equation}
	\hat{\bm{X}}_{CBoW} = \text{softmax}(\bm{R})
\end{equation}

\subsubsection{Loss Function}
Parameters of LCD module are learned through the optimization of the variational lower bound on the marginal log likelihood of features:
\begin{equation}     
	\log p_{\varphi}(\bm{X}) = \mathbb{E}_{\bm{Z} \sim q_{\phi}(\bm{Z}|\bm{X})} \left[ \log p_{\varphi}(\bm{X}|\bm{Z}) \right] - \text{KL}(q_{\phi}(\bm{Z}|\bm{X}) \| p(\bm{Z}))
	\label{equation: reconstrution loss}
\end{equation}
where $\phi$ and $\varphi$ are parameters of the encoder and decoder, respectively. The KL divergence term ensures that the distribution $q_{\phi}(\bm{Z}|\bm{X})$ approaches the prior probability $p(\bm{Z})$, while $p_{\varphi}(\bm{X}|\bm{Z})$ represents the decoding process.

\subsection{Generative Framework}
\label{subsection: Generative Framework}

\subsubsection{Encoder}
The generative framework (refer to \Cref{fig: architecture}) employs a T5 encoder \cite{RaffelSRLNMZLL20} for input encoding. Given a comment $X = [x_1,\cdots ,x_N]$, the encoder first converts the token sequence into a high-dimensional vector representation $\bm{E} \in \mathbb{R}^{N \times dim}$ using an internal embedding layer, where \(N\) is the sequence length, and \textit{dim} is the dimension of the representation.

Subsequently, $\bm{E}$ is passed through the T5 encoder which comprises multiple layers of Transformer \cite{vaswani2017attention} structure, to calculate context-aware representations:
\begin{equation}
	\bm{H}^{enc} = \text{T5Encoder}(\bm{E}) 
\end{equation}
where $\bm{H}^{enc} \in \mathbb{R}^{N \times dim}$.

\subsubsection{Category-Text Attention}
After encoding the input sequence into representation $\bm{H}^{enc}$, the model computes attention score of the latent category distribution over the input sequence, to establish the relationship between them. The attention score is calculated as follows:
\begin{equation}
	\bm{e}_j = \frac{\text{dot}(\bm{R}^{lcd}, \bm{H}_j^{enc})}{\sqrt{dim}}
\end{equation}
where $\text{dot}(\cdot)$ is the dot product, and $j \in [1, N]$. Following this, $\bm{e}$ is fed to a softmax function to calculate the attention weights:
\begin{equation}
	\bm{a} = \text{softmax}(\bm{e})
\end{equation}
$\bm{a}$ is then applied as weights to the representation of the input sequence to obtain the output vector $\bm{V}$ of the encoder:
\begin{equation}
	\bm{V}_j = \bm{a}_j \bm{H}_j^{enc}
\end{equation}
The vector $\bm{V}$, representing the input text and incorporating LCD information, is then fed into the decoder to provide contextual information for decoding.

\subsubsection{Decoder}
The model employs a T5 decoder \cite{RaffelSRLNMZLL20} to sequentially generate the output sequence $Y=[y_1, \ldots, y_i, \ldots]$ in an auto-regressive manner. When predicting the output at step $i$, the decoder uses the context vector $\bm{V}$ and the previous outputs $y_{<i}$ as inputs to calculate the current hidden state of the decoder:

\begin{equation}
	\bm{H}_i^{dec} = \text{T5Decoder}(\bm{V}, y_{<i})
\end{equation}
Here, each $\text{T5Decoder}(\cdot)$ layer comprises a Transformer structure and a cross-attention mechanism \cite{vaswani2017attention} for enhancing the representation of $\bm{H}_i^{dec}$. The probability distribution over target vocabulary at step $i$ is calculated as follows:
\begin{equation}
	P(y_i|y_{<i},\bm{V}) = \text{Softmax}(\bm{W}\bm{H}_i^{dec} + \bm{b})
\end{equation}
where $\bm{W} \in \mathbb{R}^{|\mathcal{V}| \times dim}$, $\mathcal{V}$ is the target vocabulary, and $\bm{b}$ is a bias vector.

The decoder repeats this procedure until it encounters the end token ``$\langle eos \rangle$'' or reaches the maximum specified output length.

\subsubsection{Loss Function}
The loss function of the generative framework is defined as the cross-entropy loss between the generated sequence and the target sequence:
\begin{equation}
	L(\theta) = -\sum_{i=1}^{M} \log P(y_i|y_{<i}, \bm{V})
	\label{equation: generative loss}
\end{equation}
where $\theta$ represents the parameters of the generative framework, and $M$ is the length of the target sequence.

\subsection{Constrained Decoding}

\subsubsection{Target Sequence Pattern Knowledge}
In generative models, a commonly used decoding approach is greedy decoding \cite{0001LXHTL0LC20}, where the model calculates the probability distribution over the entire vocabulary at each decoding step, and selects the word with the highest predicted probability as the output.

However, this decoding strategy may lead to invalid structure patterns. 
For instance, 
1) due to the limited number of category types, greedy decoding using the entire vocabulary may yield invalid categories,
2) since each category corresponds to distinct subcategories, greedy decoding might generate subcategories that do not align with that category,
3) as sentiment polarity is finite, greedy decoding could result in invalid sentiment polarities,
4) incomplete sequence structures, such as the failure to generate pairs of brackets, may occur. 
Following Lu et al. \cite{0001LXHTL0LC20}, information regarding the structural patterns of the target sequence, which can be used to regulate its generation process, is referred to as target sequence pattern knowledge in this paper.

\subsubsection{Constrained Decoding Strategy}
\begin{figure}[htbp]
	\centering
	\begin{subfigure}{0.63\textwidth}
		\centering
		\setlength{\FrameSep}{0pt}%
		\setlength{\abovedisplayskip}{0pt}%
		\setlength{\abovedisplayshortskip}{0pt}%
		\setlength{\belowdisplayskip}{0pt}%
		\setlength{\belowdisplayshortskip}{0pt}%
		\adjustbox{max width=0.8\textwidth}{
			\begin{tikzpicture}
				\tikzset{
					grow'=right,
					execute at begin node=\strut,
					level 1/.style={level distance=18pt},
					level 2/.style={level distance=48pt},
					sibling distance=13pt,
				}
				
				\node {$\mathcal{C}_{1}$}
				child {node[anchor=west] {HARDWARE ($\mathcal{C}_{2}$)}
					child {node[anchor=west] {PORTABILITY}}
					child {node[anchor=west] {$\cdots$}}
					child {node[anchor=west] {USABILITY}}
				}
				child {node[anchor=west] {CPU}}
				child {node[anchor=west] {$\cdots$}}
				child {node[anchor=west] {KEYBOARD ($\mathcal{C}_{2}$)}{
						child {node[anchor=west] {PRICE}}
						child {node[anchor=west] {$\cdots$}}
						child {node[anchor=west] {QUALITY}}}
				};
			\end{tikzpicture}
		}
		\captionsetup{skip=0pt} 
		\caption{Trie of category $\mathcal{C}_{1}$ and respective subcategory $\mathcal{C}_{2}$}
		\label{fig: Trie for category}
	\end{subfigure}
	\hfill
	\begin{subfigure}{0.33\textwidth}
		\centering
		\adjustbox{max width=0.7\textwidth}{
			\begin{tikzpicture}
				\tikzset{grow'=right,level distance=18pt, sibling distance=16pt,}
				\tikzset{execute at begin node=\strut}
				\node[anchor=west] {$\mathcal{S}$}
				child {node[anchor=west] {Positive}}
				child {node[anchor=west] {Negative}}
				child {node[anchor=west] {Neutral}}				
				;
			\end{tikzpicture}
		}
		\captionsetup{skip=0pt} 
		\caption{Trie of sentiment types $\mathcal{S}$}
		\label{fig: Trie for sentiment}
	\end{subfigure}
	
	\vspace{0.3cm} 
	
	\begin{subfigure}{0.90\textwidth}
		\centering
		\setlength{\FrameSep}{0pt}%
		\setlength{\abovedisplayskip}{0pt}%
		\setlength{\abovedisplayshortskip}{0pt}%
		\setlength{\belowdisplayskip}{0pt}%
		\setlength{\belowdisplayshortskip}{0pt}%
		\adjustbox{max width=0.8\textwidth}{
			\begin{tikzpicture}\tikzset{
					grow'=right,
					execute at begin node=\strut,
					level 1/.style={level distance=24pt},
					level 2/.style={level distance=24pt},
					sibling distance=16pt,
				}
				\node {$\langle bos \rangle $}
				child {node {$[$}
					child {node {$\langle$}
						child {node {$\mathcal{A}$}
							child {node {$\mathcal{C}_{1}$}
								child {node {$\mathcal{C}_{2}$}
									child {node {$\mathcal{O}$}
										child {node {$\mathcal{S}$}
											child {node {$ \rangle$}
												child {node {$ \langle $}
													child {node {$\mathcal{A}$}
														child {node {$\cdots$}}
													}
												}
												child {node {$]$}
													child {node {$\langle eos \rangle $}}
												}
											}
										}
									}
								}
							}
						}
					}
					child {node {$]$}
						child {node {$\langle eos \rangle $}
						}
					}
				};
			\end{tikzpicture}
		}
		\captionsetup{skip=0pt} 
		\caption{Trie of linearized sentiment quadruples}
		\label{fig: Trie for quadruple}
	\end{subfigure}
	\caption{Prefix tree (trie) for constrained decoding. $\mathcal{C}_1$ and $\mathcal{C}_2$ denote the set of category and respective subcategory, $\mathcal{S}$ indicates the set of sentiment, $\mathcal{A}$ and $\mathcal{O}$ denote the aspect and opinion text span, respectively.}
\end{figure}
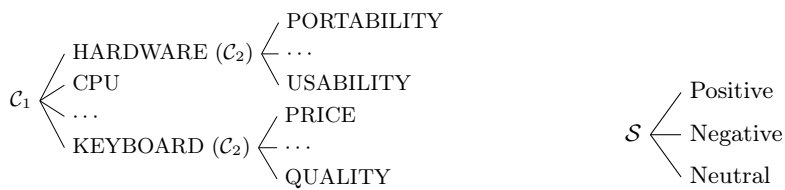
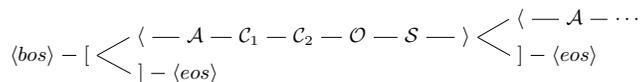

Pattern knowledge of the target sequence can constrain the decoding process, reducing the search space and enhancing target generation. 
To leverage this knowledge, this study employs a trie-based constrained decoding strategy \cite{ChenBHK20}. 
A trie, also referred to as a prefix tree, is a type of $k$-ary search tree that serves as a data structure for locating specific keys within a set \cite{Fredkin60}.
With this structure, constrained decoding strategy dynamically selects and prunes candidate vocabulary based on the current generation state. The candidate vocabulary $\mathcal{V}'$ includes the following three types:
\begin{enumerate}
	\item Structure pattern: category set $\mathcal{C}_1$ (see \Cref{fig: Trie for category}), their respective subcategory set $\mathcal{C}_2$ (see \Cref{fig: Trie for category}), and sentiment polarities set $\mathcal{S}$ (see \Cref{fig: Trie for sentiment}).
	\item Text span: aspects $\mathcal{A}$ and opinions $\mathcal{O}$, representing text spans in the input.
	\item Structure indicators: ``$[$'',  ``$]$'', ``$\langle$'', and ``$\rangle $'' (see \Cref{fig: Trie for quadruple}), utilized to combine structure patterns and text spans.
\end{enumerate}

\Cref{fig: Trie for quadruple} depicts constrained decoding using a trie of linearized sentiment quadruples. Starting with the initial token ``$\langle bos \rangle$'', candidate vocabulary $\mathcal{V}'$ at each step is confined to child nodes stemming from the previously generated node. 
For example, after generating the sequence ``$\langle bos \rangle$ $[$'', $\mathcal{V}'$ is limited to the child node set \{``$\langle$'', ``$]$''\} of node ``$[$''.



When generating the category, decoding involves searching the trie structure of category $\mathcal{C}_1$ (\Cref{fig: Trie for category}). Similarly, when generating the subcategory, decoding searches the corresponding subtree, constraining the search space based on the category-subcategory dependency. For example, when generating the subcategory of category ``\textit{HARDWARE}'', the candidate vocabulary $\mathcal{V}'$ is the child node set of ``\textit{HARDWARE}'', namely, \{``\textit{PORTABILITY}'', $\cdots$, ``\textit{USABILITY}''\}.

To generate sentiment polarity, decoding is confined to searching the sentiment polarity trie $\mathcal{S}$ (\Cref{fig: Trie for sentiment}). For aspect words and opinion words generation, the candidate vocabulary $\mathcal{V}'$ is the entire T5 vocabulary. This generation process repeats until the termination token ``$\langle eos \rangle$'' is produced. Finally, the output sequence is converted into quadruples to serve as the final extraction results.

\section{Experiments}
\subsection{Datasets}
\label{section: Datasets}
Two datasets, Restaurant-ACOS \cite{CaiXY20} and Laptop-ACOS \cite{CaiXY20}, are used to evaluate model performance. Each sample comprises a review text and annotated sentiment quadruples, as summarized in \Cref{tab: Dataset Samples and Quadruple Statistics}.


These datasets also introduce the concept of \textit{implicit} elements, where \textit{aspects} or \textit{opinions} within the review are considered \textit{implicit} if they are not explicitly stated. 
For example, in the review ``\textit{apparently apple isn't even trying anymore}'', the annotated quadruple $\langle$\textit{apple}, \textit{COMPANY}\textit{\#}\textit{GENERAL}, \textit{Null}, \textit{NEG}$\rangle$ indicates that the \textit{opinion} towards the \textit{aspect} ``\textit{apple}'' is not explicitly stated in the review and is considered \textit{implicit}, labeled as ``\textit{Null}'' in the annotation.


The statistics of implicit elements is shown in \Cref{tab: Statistics of Implicit Elements in the Datasets}. Here, EA\&EO represents quadruples with only explicit aspects and opinions, IA\&EO indicates those with implicit aspects and explicit opinions, EA\&IO covers quadruples with explicit aspects and implicit opinions, and IA\&IO denotes quadruples with only implicit aspects and opinions.

\begin{table}[htbp]
	\centering
	\begin{minipage}[t]{.50\linewidth}
		\centering
		\setlength{\tabcolsep}{2.0pt} 
		\caption{Statistics of Samples and Quadruples}
		\begin{adjustbox}{max width=0.99\textwidth}
                \fontsize{8.5}{10.5}\selectfont   
			\begin{tabular}{llcc} 
				\toprule
				&       & Samples & Quadruples  \\ 
				\cmidrule(r){1-4}
				\multirow{4}{*}{R-ACOS} & Train & 1530     & 2484        \\
				& Dev   & 171      & 261         \\
				& Test  & 583      & 916         \\
				& Total & 2284     & 3661        \\ 
				\cmidrule(r){1-4}
				\multirow{4}{*}{L-ACOS}     & Train & 2934     & 4192        \\
				& Dev   & 326      & 440         \\
				& Test  & 816      & 1161        \\
				& Total & 4076     & 5773        \\
				\bottomrule
			\end{tabular}
		\end{adjustbox}	
		\label{tab: Dataset Samples and Quadruple Statistics}
	\end{minipage}%
	\hfill
	\begin{minipage}[t]{.45\linewidth}
		\centering
		\setlength{\tabcolsep}{2.0pt} 
		\caption{Statistics of Implicit Elements in the Datasets}
		\begin{adjustbox}{max width=0.99\textwidth}
                \fontsize{8.5}{10.5}\selectfont
			\begin{tabular}{ccc} 
				\toprule
				& R-ACOS & L-ACOS  \\ 
				\cmidrule(r){1-3}
				EA\&EO  & 2429            & 3269         \\
				IA\&EO  & 530             & 910          \\
				EA\&IO  & 350             & 1237         \\
				IA\&IO  & 349             & 342          \\ 
				\cmidrule(r){1-3}
				Total   & 3658            & 5758         \\
				\bottomrule
			\end{tabular}
		\end{adjustbox}
			
		\label{tab: Statistics of Implicit Elements in the Datasets}
	\end{minipage}
\end{table}

\subsection{Experimental Setup}

A GeForce RTX 3090 (24GB) was utilized for computation. T5-base model was employed with a training batch size of 16, and encoding and hidden layer dimensions set to 768. The AdamW optimizer was employed with a learning rate of 2e-5, and dropout was applied at a rate of 0.2. Experiments were initialized five times randomly, and the average result was considered the final outcome.

While it was possible to train the model solely with the generation framework loss (\Cref{equation: generative loss}), practical experiments revealed slow convergence and poor performance. To tackle this issue, the LCD module was initially trained with the reconstruction loss (\Cref{equation: reconstrution loss}) until convergence. Subsequently, the entire model was trained using the generation loss (\Cref{equation: generative loss}). Once the generation loss approached convergence, further training was applied to the LCD module until convergence. This process was repeated until the model demonstrated improved performance.

Evaluation employed Precision, Recall, and F1 score, with a quadruple being deemed correctly predicted only if all its four elements were predicted accurately.

\subsection{Baselines}
The subsequent models were employed for comparative analysis:

Extractive:
\textbf{DP-ACOS} \cite{CaiXY20}: Utilizes the DP algorithm to extract sentiment triplets $ \langle a,o,s \rangle $ and then determines the category of each triplet.
\textbf{JET-ACOS} \cite{CaiXY20}: Employs JET \cite{xu2020position} to extract triplets $ \langle a,o,s \rangle $ and then predicts category of each triplet.
\textbf{TB-ACOS} \cite{CaiXY20}: Utilizes TAS-BERT \cite{wan2020target} for aspect-opinion joint extraction based on category-sentiment, then filters out invalid pairs.
\textbf{EC-ACOS} \cite{CaiXY20}: Performs aspect-opinion joint extraction, then predicts category-sentiment pairs.
\textbf{SGTS} \cite{ZhuBXLZK23}: Captures high-dimensional features of sentences, then uses grid labeling scheme and its decoding method for extraction.

Generative:
\textbf{BARTABSA} \cite{yan-etal-2021-unified}: A model based on BART \cite{lewis-etal-2020-bart}, generates sentiment quadruples by predicting the indices of target words and sentiment categories.
\textbf{GAS} \cite{Zhang0DBL20}: Using T5, GAS models sentiment analysis and various subtasks as generative tasks.
\textbf{Paraphrase} \cite{ZhangD0YBL21}: Transforms the quadruple extraction task into a paraphrase generation process using predefined templates.
\textbf{Seq2Path} \cite{MaoSYZC22}: Formalizes the generated content as a tree, with sentiment tuples representing the paths of the tree.

\section{Experimental Results and Analyses}
\subsection{Main Results}

\Cref{tab: main results} presents the experimental results of the proposed model and baseline models on two datasets. The results highlight the superiority of generative models over extractive models. This advantage can be attributed primarily to two factors: 1) treating the extraction task as an end-to-end sequence generation problem reduces error propagation; 2) uniformly modeling the generation of each quadruple element facilitates more effective utilization of semantic knowledge in pretrained models, enhancing semantic sharing and decision interaction.

Furthermore, the results in \Cref{tab: main results} suggest that the proposed model, utilizing LCD and CD, outperforms the comparing generative models on both datasets. Specifically, on the Restaurant-ACOS dataset, compared to the baseline model GAS, the proposed model shows a decrease of 0.004 in Precision, but an increase of 0.038 in Recall, and an increase of 0.017 in F1. On the Laptop-ACOS dataset, the proposed model exhibits an increase of 0.031 in Precision, 0.035 in Recall, and 0.033 in F1. These results highlight the superior performance of the proposed model in sentiment quadruple extraction.

Despite a notable improvement in performance compared to the baseline models, the overall performance remains modest. This can be attributed to two main factors. 
First, the inherent complexity of the task necessitates the extraction and precise matching of four distinct types of sentiment elements, thereby increasing the difficulty of the task. 
Second, special expressions in language such as the presence of implicit elements within the text presents a significant challenge, potentially impeding the model's comprehensive semantic understanding, and consequently leading to suboptimal performance. 
The following subsection will delve into the extraction of implicit elements.

\begin{table}[htbp]
	\centering
	\caption{Main Results}
	
	\begin{adjustbox}{max width=0.8\textwidth}
            \fontsize{9}{11}\selectfont
		\begin{tabular}{p{1.6cm} p{2.4cm} p{1.1cm} p{1.1cm} p{1.1cm} p{1.1cm} p{1.1cm} p{1.0cm}} 
			\toprule
			&  & \multicolumn{3}{l}{Restaurant-ACOS} & \multicolumn{3}{l}{Laptop-ACOS} \\ 
			\cmidrule(r){3-5}\cmidrule{6-8}
			&  & P & R & F1 & P & R & F1 \\ 
			\cmidrule(r){1-2}\cmidrule(r){3-5}\cmidrule{6-8}
			\multirow{5}{*}{Extractive} & DP-ACOS & 0.347 & 0.151 & 0.210 & 0.130 & 0.006 & 0.080 \\
			& TB-ACOS & 0.263 & 0.463 & 0.335 & \textbf{0.472} & 0.192 & 0.273 \\
			& JET-ACOS & 0.598 & 0.289 & 0.390 & 0.445 & 0.163 & 0.238 \\
			& EC-ACOS & 0.385 & 0.530 & 0.446 & 0.456 & 0.295 & 0.358 \\
			& SGTS & 0.552 & 0.437 & 0.488 & 0.414 & 0.323 & 0.363 \\ 
			\cmidrule(r){1-2}\cmidrule(r){3-5}\cmidrule{6-8}
			\multirow{5}{*}{Generative} & BARTABSA & 0.566 & 0.554 & 0.560 & 0.417 & 0.405 & 0.411 \\
			& Seq2Path & N/A & N/A & 0.584 & N/A & N/A & 0.430 \\
			& Paraphrase & 0.590 & 0.591 & 0.590 & 0.418 & 0.450 & 0.433 \\
			& GAS & \textbf{0.607} & 0.585 & 0.596 & 0.416 & 0.428 & 0.422 \\
                \cdashline{2-8}
			& Proposed model & 0.603 & \textbf{0.623} & \textbf{0.613} & 0.447 & \textbf{0.463} & \textbf{0.455} \\
			\bottomrule
		\end{tabular}
	\end{adjustbox}
	\label{tab: main results}
\end{table}

\subsection{Implicit Element Extraction Results}

As noted in \Cref{section: Datasets}, both datasets include implicit elements \cite{CaiXY20}, which are not explicitly stated in the reviews. These elements necessitate a comprehensive understanding of the text and present a challenge for models, thus attracting considerable attention in recent sentiment analysis research \cite{ZhangD0YBL21,ZhuBXLZK23}.

To assess the capacity of models in extracting implicit elements, the test set was divided into four subsets: EA\&EO, IA\&EO, EA\&IO, and IA\&IO, each targeting specific type of quadruples, whose meanings were introduced in \Cref{section: Datasets}. Results in \Cref{tab: Results of the Extraction of Implicit Elements} indicate that models excel in extracting explicit elements compared to implicit ones. For instance, on the Restaurant-ACOS dataset, the proposed model achieves an F1 score of 0.661 on the EA\&EO subset, which is comprised solely of explicit elements, significantly outperforming scores of 0.544, 0.442, and 0.413 on the other three subsets containing implicit elements. Similar trends are also observed on the Laptop-ACOS dataset. These results highlight the challenge of extracting implicit elements.

Moreover, \Cref{tab: Results of the Extraction of Implicit Elements} shows that, on Restaurant-ACOS, the proposed model not only outperforms baseline models on the EA\&EO subset, but also on subsets containing implicit elements (IA\&EO and IA\&IO). 
Similarly, on Laptop-ACOS, the proposed model also outperforms on subsets comprising implicit elements (EA\&IO and IA\&IO). These findings highlight the capability of the proposed model in deep semantic comprehension and generation of implicit elements.


\begin{table}[htbp]
	\centering
	\setlength{\tabcolsep}{1.5pt} 
	\caption{Results of the Extraction of Implicit Elements (F1)}
	\begin{adjustbox}{max width=0.8\textwidth}
            \fontsize{8.5}{10.5}\selectfont
		\begin{tabular}{lllllllll} 
			\toprule
			& \multicolumn{4}{l}{Restaurant-ACOS} & \multicolumn{4}{l}{Laptop-ACOS}  \\ 
			\cmidrule(r){2-5}\cmidrule{6-9}
			& EA\&EO  & IA\&EO  & EA\&IO  & IA\&IO        & EA\&EO  & IA\&EO  & EA\&IO  & IA\&IO     \\ 
			\cmidrule(r){1-1}\cmidrule(r){2-5}\cmidrule{6-9}
			DP-ACOS               & 0.260 & N/A   & N/A   & N/A         & 0.098  & N/A   & N/A   & N/A      \\
			JET-ACOS              & 0.523 & N/A   & N/A   & N/A         & 0.357 & N/A   & N/A   & N/A      \\
			TB-ACOS         & 0.336 & 0.318 & 0.140 & 0.398       & 0.261 & 0.415 & 0.109 & 0.212    \\
			EC-ACOS & 0.450 & 0.347 & 0.239 & 0.337       & 0.354 & 0.390 & 0.168 & 0.186    \\
			SGTS                  & 0.560 & 0.396 & 0.111 & 0.296       & 0.366 & \textbf{0.525} & 0.172 & 0.146    \\
			Paraphrase            & 0.654 & 0.533 & \textbf{0.456} & N/A         & 0.457 & 0.510 & 0.330 & N/A      \\
            \cdashline{1-9}
			Proposed        & \textbf{0.661} & \textbf{0.544} & 0.442 & \textbf{0.413}       & \textbf{0.498} & 0.492 & \textbf{0.347} & \textbf{0.283}    \\
			\bottomrule
		\end{tabular}
	\end{adjustbox}
		
	\label{tab: Results of the Extraction of Implicit Elements}
\end{table}

\vspace{-1em}
\subsection{Ablation Study}

To assess the effectiveness of the proposed LCD module and CD strategy, ablation experiments were conducted on the datasets. Results in \Cref{tab: Ablation Study} indicate the impact of removing the LCD module (\textit{w/o LCD}), excluding the CD strategy (\textit{w/o CD}), and simultaneously excluding both components (\textit{w/o LCD\&CD}).

As shown in \Cref{tab: Ablation Study}, performance deteriorates when LCD and CD are excluded. Specifically, on the Restaurant-ACOS dataset, the removal of LCD alone resulted in an F1 decrease of 0.009, while excluding CD alone led to an F1 decrease of 0.012. Simultaneous removal of both components resulted in an F1 decrease of 0.014. Similar trends were also observed on the Laptop-ACOS dataset. 
These results suggest that the proposed LCD module enhances the extraction performance by capturing latent category features, while the CD strategy regulates and guides the generation process, mitigating the generation of erroneous patterns.

\begin{table}[!h]
	\centering
        \setlength{\tabcolsep}{1.5pt} 
	\caption{Ablation Study}
	\begin{adjustbox}{max width=0.7\textwidth}
                \fontsize{8.5}{10.5}\selectfont
			\begin{tabular}{lllllll} 
				\toprule
				& \multicolumn{3}{l}{Restaurant-ACOS} & \multicolumn{3}{l}{Laptop-ACOS}  \\ 
				\cmidrule(r){2-4}\cmidrule(r){5-7}
				& P     & R     & F1                  & P     & R     & F1               \\ 
				\cmidrule(r){1-1}\cmidrule(r){2-4}\cmidrule(r){5-7}
				Proposed model           & 0.603 & 0.623 & \textbf{0.613}               & 0.447 & 0.463 & \textbf{0.455}            \\
				\quad\textit{w/o LCD}                   & 0.596 & 0.613 & 0.604 (-0.009)               & 0.435 & 0.463 & 0.449 (-0.006)           \\
				\quad\textit{w/o CD}                    & 0.591 & 0.610 & 0.601 (-0.012)              & 0.433 & 0.450 & 0.441 (-0.014)           \\
				\quad\textit{w/o LCD\&CD} & 0.587 & 0.612 & 0.599 (-0.014)              & 0.426 & 0.452 & 0.438 (-0.017)           \\
				\bottomrule
			\end{tabular}
	\end{adjustbox}
		
	\label{tab: Ablation Study}
\end{table}

\subsection{Visualization of Latent Category Distribution}

\begin{figure}[!h]
	\centering
	\includegraphics[width=0.9\linewidth]{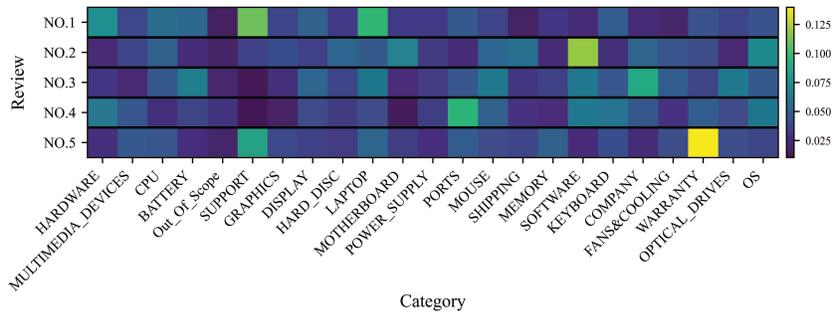}
	\caption{Visualization of Latent Category Distribution}
	\label{fig: Visualization of Latent Category Distribution}
\end{figure}

As introduced in \Cref{section: Latent Category Distribution}, the LCD module learns latent category distribution representation $\bm{Z}$ for a sample. To visually illustrate the effectiveness of this module, five samples were selected with their learned LCD representations visualized in \Cref{fig: Visualization of Latent Category Distribution}. The $x$-axis denotes the total 23 categories, while the $y$-axis indicates the sample indices. 
Each row in the figure corresponds to a LCD representation vector of a sample across categories, with color intensity reflecting the magnitude of the vector in respective category.
The five samples, along with their respective standard categories in parentheses, are as follows:
\begin{enumerate}[left=1em]
	\item \textit{the product is great, but the customer support is horrible.} (\textit{LAPTOP, SUPPORT})
	\item \textit{second issue is with scaling of the ui.} (\textit{SOFTWARE})
	\item \textit{asus, has a horrible reputation.} (\textit{COMPANY})
	\item \textit{hdmi out doesn't work right.} (\textit{PORTS})
	\item \textit{now i have to deal with warranty stuff and sending it back.} (\textit{WARRANTY})
\end{enumerate}

\vspace{-1.5em} 

\vspace{-1em} 
As depicted in \Cref{fig: Visualization of Latent Category Distribution}, the LCD module effectively captures latent category distribution information for samples, including those with complex category inclusion (sample 4) or overlap (sample 5) relationships. 
For example, while the standard category of sample 4 is ``\textit{PORTS}'', it is highly predictable as ``\textit{HARDWARE}'' or ``\textit{LAPTOP}'' due to semantic inclusion relationships, where ``\textit{PORTS}'' is semantically included by both ``\textit{HARDWARE}'' and ``\textit{LAPTOP}''. 
For sample 5, although its standard category is ``\textit{WARRANTY}'', it is highly predictable as ``\textit{SUPPORT}'' due to semantic overlap between the two categories. \Cref{fig: Visualization of Latent Category Distribution} demonstrates the accurate capture of latent category distributions by LCD.

\section{Conclusion}
This study introduces a generative model for extracting sentiment quadruples. To address challenges related to semantic inclusion and overlap among categories, the model incorporates a latent category distribution to capture the strength of the relationship between text and categories. Moreover, a constrained decoding strategy based on a trie structure is employed to exploit knowledge of target sequence patterns, thereby reducing the search space and enhancing the generation process. Experimental results demonstrate performance enhancements compared to baseline models, with ablation studies verifying the efficacy of the latent category distribution and constrained decoding strategy.
Despite advancements, performance remains moderate, primarily due to task complexity and special expressions in language, including implicit elements.
Future research could focus on enhancing feature extraction and representation, as well as improving contextual understanding, and integrating coreference resolution and background knowledge, to improve the extraction process, including implicit elements.

\begin{credits}
\subsubsection{\ackname}
This work was supported by the National Natural Science Foundation of China (No. 62176187), the National Key Research and Development Program of China (No. 2022YFB3103602, No. 2017YFC1200500), the Research Foundation of Ministry of Education of China (No. 18JZD015).

\end{credits}
%
%
%
\bibliographystyle{splncs04}
\bibliography{mybibfile}
%

\end{document}